\documentclass[journal]{IEEEtran}

\usepackage[T1]{fontenc}
\usepackage{amsmath,amsfonts}
\usepackage{algorithm}
\usepackage{algorithmic}
\usepackage{booktabs}
\usepackage{cite}
\usepackage{graphicx}

\begin{document}

\title{Multi-Appliance Non-Intrusive Load Monitoring via
Label-Preserving Aggregate Recomposition and Prediction Consistency}

\author{Jiangfeng~Liu and Yanfang~Fan%
\thanks{The authors are with the School of Electrical Engineering, Xinjiang University, Urumqi, Xinjiang 830047, China (e-mail: ljf\_cns@163.com; fyf\_xj@sina.com). \textit{(Corresponding author: Yanfang Fan.)}}}

\maketitle

\begin{abstract}
Non-intrusive load monitoring (NILM) estimates appliance power sequences from aggregate power, but models trained on source households commonly lose accuracy in unseen households.
Aggregate power also contains loads from other appliances and measurement error, so predictions may depend on the residual background that co-occurs with source-household targets.
Time-aligned submetered measurements and the additive decomposition of aggregate power expose a relation unused by window-wise supervision: an aggregate window can be recomposed by replacing only its residual background while preserving all modeled target-appliance power sequences pointwise.
We combine label-preserving aggregate recomposition with prediction consistency. Both windows receive complete power and operating-state supervision.
For each appliance, disagreement between the two power predictions is penalized only when both satisfy a fixed reliability criterion and only to the extent that it exceeds a fixed margin.
The proposed method is implemented using a multi-appliance architecture with two-stage shared-to-specific mixture-of-experts routing.
On REDD, UK-DALE, and REFIT, the proposed method lowers appliance-averaged mean absolute error relative to single-window training from 14.75 to 13.14 W, from 8.88 to 8.51 W, and from 15.83 to 14.55 W.
Label-preserving aggregate recomposition and prediction consistency are used only during training, and add no inference-time module or parameter.
\end{abstract}


\section{Introduction}
\label{sec:introduction}

\IEEEPARstart{B}{uildings} account for a substantial share of global energy demand, and residential use represents most building-sector consumption~\cite{IEA2025EnergyEfficiency}.
Information on when individual appliances operate and how much electricity they use can support residential energy management and inform demand-side management~\cite{SchirmerMporas2023Review}.
Non-intrusive load monitoring (NILM) estimates appliance power consumption from aggregate smart-meter readings without requiring individual metering of every appliance~\cite{Hart1992NILM,LinEtAl2022KTLN}.
NILM models learn this mapping during training on labeled data from source households.

\subsection{Background and Motivation}

This paper considers a multi-appliance NILM setting in which a single model jointly estimates the power sequences of several target appliances from the same aggregate power sequence~\cite{DashSahoo2024MultiTask}.
NILM models can show performance degradation when applied to previously unseen households~\cite{LinEtAl2022KTLN}.
This degradation may arise from differences across households in the power consumption of both the target appliances and the other appliances.
Even appliances of the same type can exhibit different power characteristics across brands or models~\cite{RafiqEtAl2021Augmentation}, and household usage habits can change when and for how long the target appliances operate~\cite{LinEtAl2022KTLN,LuoEtAl2024MetricMeta}.
The aggregate also contains power consumption from appliances outside the target set, and this contribution can vary across households~\cite{XiongEtAl2024MATNilm,ChenEtAl2023EndCloud}.

For a specified set of target appliances, we define the residual background as the difference between the aggregate power and the sum of the target-appliance power sequences.
It comprises consumption from all other appliances together with measurement error.
In the training data, patterns in the residual background may correlate with target-appliance operation, and a model may exploit these correlations~\cite{LiEtAl2026CRMAT}.
If these correlations no longer hold in other households, predictions that rely on them may become less accurate~\cite{HeinzeDemlMeinshausen2021CoRe}.
During training, we therefore seek to limit how much the model relies on these correlations.

\subsection{Related Work and Research Gap}

Studies in computer vision have examined models' reliance on background cues by deliberately altering non-target context~\cite{XiaoEtAl2021Backgrounds,RoderSchweighofer2026AutoBackSwap}.
SwapMix~\cite{GuptaEtAl2022SwapMix} swaps the features of question-irrelevant context objects identified using ground-truth question reasoning steps, whereas Counterfactual Generative Networks~\cite{SauerGeiger2021CGN} construct training images by recomposing disentangled visual factors.
Although these constructions differ, they share the idea of varying factors that should not determine the target label while retaining task-relevant semantics.
Both constructions, however, depend on annotations or on a generative decomposition, and they preserve categorical target semantics rather than dense, pointwise regression targets.

In NILM, several studies constrain learned representations through relations defined over modified or paired observations.
CR-MAT mixes the frequency magnitudes of an aggregate window with those from another household while retaining the original phase, and penalizes the representation change caused by this mixing~\cite{LiEtAl2026CRMAT}.
Han \emph{et al.} form positive pairs from matching timestamps across augmented contexts and construct hard negatives by mixing features in the embedding space~\cite{HanEtAl2026HardNegative}.
DisCoV instead separates positive from negative samples according to whether a given appliance is active, and constrains appliance-specific latent variables~\cite{OublalEtAl2024DisCoV}.
Outside NILM, CoRe provides a related grouped-observation formulation that penalizes the conditional variance of predictions within groups sharing the same target and identifier~\cite{HeinzeDemlMeinshausen2021CoRe}.

To limit reliance on residual-background correlations during training, we seek a window pair whose two windows differ only in the residual background, thereby leaving every modeled target-appliance power sequence pointwise unchanged.
The reviewed approaches do not provide such a pair.
In the supervised multi-appliance setting, however, such a pair can be constructed directly from time-aligned submetered measurements using the additive decomposition of aggregate power.
Because the two windows have the same targets, training can constrain the disagreement between their corresponding per-appliance power predictions.
Conventional window-wise supervision evaluates each window against its targets but does not explicitly encode the relation between the paired predictions.

\subsection{Contributions of the Present Work}

To address this gap, we propose a method that combines label-preserving aggregate recomposition with prediction consistency during training.
We develop the feature-gated layered appliance mixture-of-experts (FLAME) architecture to share aggregate-level context across appliances while forming a distinct representation for each.
We use FLAME to implement the proposed method.
The key contributions of this work are as follows:
\begin{list}{\arabic{enumi})}{
  \usecounter{enumi}
  \setlength{\leftmargin}{0pt}
  \setlength{\rightmargin}{0pt}
  \setlength{\labelwidth}{1.5em}
  \setlength{\labelsep}{0.5em}
  \setlength{\itemindent}{3em}
  \setlength{\listparindent}{0pt}
  \setlength{\parsep}{0pt}
  \setlength{\itemsep}{0pt}
  \setlength{\topsep}{0pt}\setlength{\partopsep}{0pt}
}
  \item \emph{Label-Preserving Aggregate Recomposition and Prediction Consistency:}
  During supervised multi-appliance training, the proposed method recomposes an aggregate window so that only its residual background changes, leaving every modeled target-appliance power sequence and its operating-state targets pointwise unchanged.
  Both windows of the resulting pair receive the same complete task supervision.
  For each appliance, the method additionally constrains disagreement between the two windows' power predictions only when both predictions are accurate enough, without requiring them to be identical.
  \item \emph{FLAME Multi-Appliance Architecture:}
  FLAME treats each appliance as a task and organizes prediction through two-stage shared-to-specific expert routing.
  It first aggregates responses from shared and appliance-specific experts into a single shared representation and then refines that representation into one per appliance for joint power and operating-state prediction.
\end{list}

The remainder of this paper is organized as follows.
Section~\ref{sec:problem-formulation} states the multi-appliance NILM problem and defines the residual background.
Section~\ref{sec:proposed-method} presents the FLAME architecture, two residual-based window constructions, prediction consistency, and the overall training objective.
Section~\ref{sec:experiments-results} reports the experimental setup and results, and Section~\ref{sec:conclusion} concludes the paper.

\section{Problem Statement}
\label{sec:problem-formulation}

\begin{figure*}[t]
  \centering
  \includegraphics[width=\textwidth]{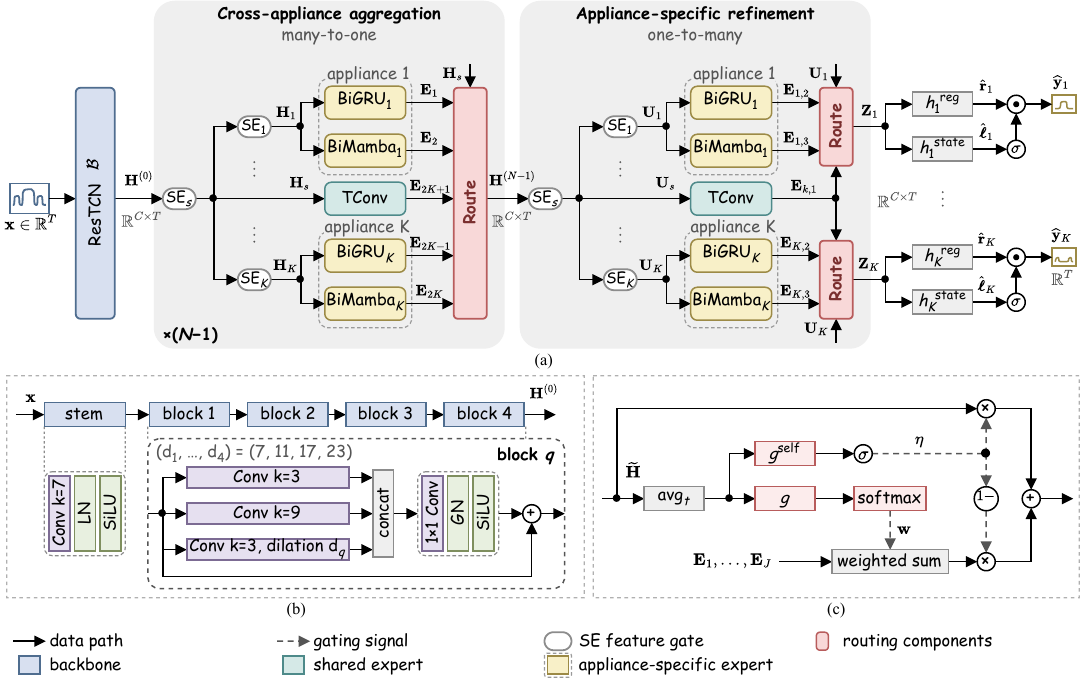}
  \caption{FLAME architecture.
  (a)~Forward topology from the backbone output $\mathbf{H}^{(0)}$ through the many-to-one aggregation levels and the one-to-many refinement level to the appliance representations $\mathbf{Z}_{k}$ and their prediction heads; rows $k=1$ and $k=K$ stand for all $K$ appliances, and line styles and colors are defined in the legend (SE: squeeze-and-excitation).
  (b)~The ResTCN backbone of~\eqref{eq:flame-backbone}, with the stem and one residual block expanded (LN: layer normalization, GN: group normalization).
  (c)~The routing operator $\mathrm{Route}(\cdot)$ of~\eqref{eq:flame-routing} and \eqref{eq:flame-routing-weights}, used with $J=2K{+}1$ in~\eqref{eq:flame-stage1} and $J=3$ in~\eqref{eq:flame-stage2}.}
  \label{fig:flame-architecture}
\end{figure*}

We consider multi-appliance non-intrusive load monitoring (NILM), where the objective is to estimate the power sequences of $K$ target appliances from aggregate active-power measurements.
Let $\mathbf{x}_i\in\mathbb{R}^{T}$ denote the $i$-th aggregate power window of length $T$, and let $\mathbf{y}_{i,k}\in\mathbb{R}^{T}$ denote the time-aligned power sequence of target appliance $k$, $k=1,\ldots,K$.
We stack the target-appliance power sequences as $\mathbf{Y}_i=[\mathbf{y}_{i,1},\ldots,\mathbf{y}_{i,K}]^{\top}\in\mathbb{R}^{K\times T}$.

The disaggregation mapping is
\begin{equation}
  \widehat{\mathbf{Y}}_i
  =F_{\theta}(\mathbf{x}_i),
  \qquad
  F_{\theta}:\mathbb{R}^{T}\rightarrow\mathbb{R}^{K\times T}
  \label{eq:disaggregation-mapping}
\end{equation}
where $\theta$ denotes the learnable parameters.
The model is trained only on labeled windows from source households and evaluated on unseen households that contribute no training data.

The aggregate window is written as
\begin{equation}
  \mathbf{x}_i
  =\sum_{k=1}^{K}\mathbf{y}_{i,k}+\mathbf{b}_i,
  \qquad
  \mathbf{b}_i
  =\mathbf{x}_i-\sum_{k=1}^{K}\mathbf{y}_{i,k}
  \label{eq:additive-observation}
\end{equation}
where $\mathbf{b}_i\in\mathbb{R}^{T}$ is the residual background defined by~\eqref{eq:additive-observation}, collecting aggregate power not represented by the target-appliance power sequences together with residual measurement and preprocessing discrepancies.

For fixed target-appliance power sequences $\mathbf{Y}_i$, the residual background can still vary, so the additive observation model admits many aggregate windows for the same targets.
It therefore permits a label-preserving aggregate recomposition that leaves $\mathbf{Y}_i$ exactly unchanged, with label preservation guaranteed by the construction itself.
The additive form also permits a complementary construction: with the residual background held fixed, substituting a target-appliance power sequence yields another additively consistent aggregate window, with its targets recomputed from the substituted sequence.
That construction produces additional training windows; only the direction that holds $\mathbf{Y}_i$ fixed yields two windows carrying identical targets.
This label-preserving relation arises from the additive decomposition and the availability of time-aligned submetered measurements, and is therefore a property of the labeled problem setting rather than of any particular training method.
Section~\ref{sec:proposed-method} describes how the proposed method uses this relation to construct anchor--recomposed pairs and selectively constrain their per-appliance power predictions.

\section{Proposed Method}
\label{sec:proposed-method}

In this section, we present the proposed method for multi-appliance NILM under window-wise supervision.
We first develop the FLAME architecture that produces one power and operating-state prediction per appliance, then define the two window constructions it is trained on, then define prediction consistency on the resulting outputs, and finally state the overall training objective.

\subsection{FLAME Multi-Appliance Architecture}
\label{sec:flame-architecture}

FLAME is designed to share aggregate-level context across appliances while forming a distinct representation for each.
It does so within a single model through two-stage shared-to-specific expert routing, illustrated in Fig.~\ref{fig:flame-architecture}(a).
A backbone extracts the initial shared representation.
The many-to-one aggregation stage routes the responses of all appliance branches into a single shared representation, and the one-to-many refinement stage expands it back into $K$ appliance views, each refined into an appliance representation from which the power and operating-state predictions are produced.
The two stages are ordered so that information is pooled across appliances before it is specialized.
The design builds on the multi-gate expert-routing paradigm of multi-gate mixture-of-experts (MMoE)~\cite{MaEtAl2018MMoE} and the shared/specific expert organization of progressive layered extraction (PLE)~\cite{TangEtAl2020PLE}.
Throughout this subsection the index $k$ ranges over the $K$ appliances, each carrying a power and an operating-state output, and the window index is omitted.

A residual temporal-convolutional backbone (ResTCN) first maps the aggregate to the initial shared representation,
\begin{equation}
  \mathbf{H}^{(0)}=\mathcal{B}(\mathbf{x})\in\mathbb{R}^{C\times T}
  \label{eq:flame-backbone}
\end{equation}
where $\mathcal{B}$ denotes the backbone, $C$ the feature-channel dimension, and the superscript indexes routing levels.
Its stem and multi-scale residual blocks, shown in Fig.~\ref{fig:flame-architecture}(b), combine local and dilated temporal convolutions following the sequence-modeling principle of residual temporal convolutional networks~\cite{BaiEtAl2018TCN}.

Both stages use the same routing form with separately parameterized gates and different candidate sets.
For a routed branch with feature-gated view $\widetilde{\mathbf{H}}\in\mathbb{R}^{C\times T}$ and $J$ candidate expert responses $\mathbf{E}_1,\ldots,\mathbf{E}_J$ of the same shape, we write $\mathrm{Route}(\cdot)$ for the gated mixing operation of Fig.~\ref{fig:flame-architecture}(c),
\begin{equation}
  \mathrm{Route}\bigl(\widetilde{\mathbf{H}};\mathbf{E}_1,\ldots,\mathbf{E}_J\bigr)
    =\eta\,\widetilde{\mathbf{H}}+(1-\eta)\sum_{m=1}^{J}w_m\,\mathbf{E}_m
  \label{eq:flame-routing}
\end{equation}
where the mixing weights $\mathbf{w}=[w_{1},\ldots,w_{J}]^{\top}$ are produced by a routing gate and the interpolation coefficient $\eta$ by a self-gate, both reading the temporally pooled view,
\begin{equation}
  \mathbf{w}=\operatorname{softmax}\!\left(
    g(\operatorname{avg}_{t}\widetilde{\mathbf{H}})\right),
  \qquad
  \eta=\sigma\!\left(
    g^{\mathrm{self}}(\operatorname{avg}_{t}\widetilde{\mathbf{H}})\right)
  \label{eq:flame-routing-weights}
\end{equation}
Here $\operatorname{avg}_{t}$ averages over time, $\sigma$ is the logistic sigmoid, and $g$ and $g^{\mathrm{self}}$ are multilayer perceptrons (MLPs) that output $J$ routing logits and a single scalar, respectively.
The routing weights form a sample-wise dense softmax, so every candidate response enters the mixture with a continuous weight.
The scalar self-gate keeps a direct path from the branch's own gated view, so the routed output interpolates between that view and the expert mixture.
The two gates thus make separate decisions: how the candidate responses are composed, and how far the routed output departs from the branch's own view.
The level index is suppressed on intermediate views and expert responses within a level, and the same operator names are reused across levels and stages for independently parameterized modules.
The configuration of these modules used in the experiments is given in Section~\ref{sec:cbc-training-setup}.

\subsubsection{Cross-Appliance Aggregation}
\label{sec:flame-stage1}

Let $N$ denote the total number of routing levels: the first $N-1$ perform cross-appliance aggregation and the final one performs appliance-specific refinement.
Aggregation level $j\in\{1,\ldots,N-1\}$ takes the shared representation $\mathbf{H}^{(j-1)}$ produced by the preceding level, the first level taking $\mathbf{H}^{(0)}$ from the backbone.
Squeeze-and-excitation (SE) feature gates~\cite{HuEtAl2018SENet} first form a shared view, marked by the subscript $s$, and from it one appliance-conditioned view per appliance, rebuilt from the current shared representation at every level,
\begin{equation}
  \mathbf{H}_{s}=\mathrm{SE}_{s}(\mathbf{H}^{(j-1)}),
  \qquad
  \mathbf{H}_{k}=\mathrm{SE}_{k}(\mathbf{H}_{s})
  \label{eq:flame-stage1-views}
\end{equation}
Because each gate pools its input over time and rescales the feature channels, conditioning leaves the temporal axis untouched and every view keeps the $C\times T$ interface of the shared representation.
All experts preserve the $C\times T$ shape.
A shared temporal-convolution expert, $\mathrm{TConv}$, is evaluated on the shared view; in both stages it gives the router one response computed from the shared view alone, so aggregate-level context that no appliance gate has reshaped remains a routing candidate.
Two heterogeneous experts are evaluated on each appliance view, consistent with the heterogeneous-expert design of~\cite{LiangEtAl2026HMMOE}: a bidirectional gated recurrent unit (BiGRU) expert~\cite{ChoEtAl2014GRU,SchusterPaliwal1997BRNN} and a bidirectional Mamba (BiMamba) expert based on input-dependent selective state-space recurrence~\cite{GuDao2023Mamba}.
The two experts present the router with responses of distinct temporal inductive bias computed from the same appliance view.
The shared router mixes all candidates with the shared view,
\begin{equation}
  \begin{gathered}
    \mathbf{E}_{2k-1}=\mathrm{BiGRU}_{k}(\mathbf{H}_{k}),
    \qquad
    \mathbf{E}_{2k}=\mathrm{BiMamba}_{k}(\mathbf{H}_{k}),\\
    \mathbf{E}_{2K+1}=\mathrm{TConv}(\mathbf{H}_{s}),\\
    \mathbf{H}^{(j)}=\mathrm{Route}\bigl(
      \mathbf{H}_{s};\,\mathbf{E}_{1},\ldots,\mathbf{E}_{2K+1}\bigr),
    \qquad j=1,\ldots,N-1
  \end{gathered}
  \label{eq:flame-stage1}
\end{equation}
The updated shared representation $\mathbf{H}^{(j)}$ is the only quantity passed to the next level.

\subsubsection{Appliance-Specific Refinement}
\label{sec:flame-stage2}

This stage, the final routing level, reverses the flow to one-to-many, starting from the single shared representation $\mathbf{H}^{(N-1)}$; its views are written $\mathbf{U}$ to mark the change of stage,
\begin{equation}
  \mathbf{U}_{s}=\mathrm{SE}_{s}(\mathbf{H}^{(N-1)}),
  \qquad
  \mathbf{U}_{k}=\mathrm{SE}_{k}(\mathbf{U}_{s})
  \label{eq:flame-stage2-views}
\end{equation}
Each appliance now owns an independent three-way router whose candidate set pairs the shared response, common to all appliances, with that appliance's own two expert responses, giving the appliance representation
\begin{equation}
  \begin{gathered}
    \mathbf{E}_{k,1}=\mathrm{TConv}(\mathbf{U}_{s}),
    \qquad
    \mathbf{E}_{k,2}=\mathrm{BiGRU}_{k}(\mathbf{U}_{k}),\\
    \mathbf{E}_{k,3}=\mathrm{BiMamba}_{k}(\mathbf{U}_{k}),\\
    \mathbf{Z}_{k}=\mathrm{Route}\bigl(
      \mathbf{U}_{k};\,\mathbf{E}_{k,1},\mathbf{E}_{k,2},\mathbf{E}_{k,3}\bigr)
  \end{gathered}
  \label{eq:flame-stage2}
\end{equation}

Two prediction heads map each appliance representation $\mathbf{Z}_{k}$ to predictions: a regression head $h_k^{\mathrm{reg}}$ produces the regression sequence $\widehat{\mathbf{r}}_{k}$, a state-classification head $h_k^{\mathrm{state}}$ produces the state-logit sequence $\widehat{\boldsymbol{\ell}}_{k}$, and one deterministic combination yields the gated power,
\begin{equation}
  \begin{gathered}
    \widehat{\mathbf{r}}_{k}=h_k^{\mathrm{reg}}(\mathbf{Z}_{k}),
    \qquad
    \widehat{\boldsymbol{\ell}}_{k}=h_k^{\mathrm{state}}(\mathbf{Z}_{k}),\\
    \widehat{\mathbf{y}}_{k}=\widehat{\mathbf{r}}_{k}\odot
      \sigma(\widehat{\boldsymbol{\ell}}_{k})
  \end{gathered}
  \label{eq:representation-interface}
\end{equation}
where $\odot$ denotes elementwise multiplication; the gated power is thus a parameter-free function of the two head outputs.
Stacking the $K$ gated outputs gives $\widehat{\mathbf{Y}}=[\widehat{\mathbf{y}}_{1},\ldots,\widehat{\mathbf{y}}_{K}]^{\top}$, the output of $F_{\theta}$ in~\eqref{eq:disaggregation-mapping}.
For training, each appliance is additionally associated with a binary operating-state target $\mathbf{s}_{k}\in\{0,1\}^{T}$, obtained by applying the configured operating-state threshold to $\mathbf{y}_{k}$ in the raw-power domain, and these targets are stacked as $\mathbf{S}=[\mathbf{s}_{1},\ldots,\mathbf{s}_{K}]^{\top}\in\{0,1\}^{K\times T}$.
They supervise the state heads in the training objective of Section~\ref{sec:joint-optimization}.

\subsection{Residual-Based Window Construction}
\label{sec:window-construction}

\begin{figure*}[t]
  \centering
  \includegraphics[width=\textwidth]{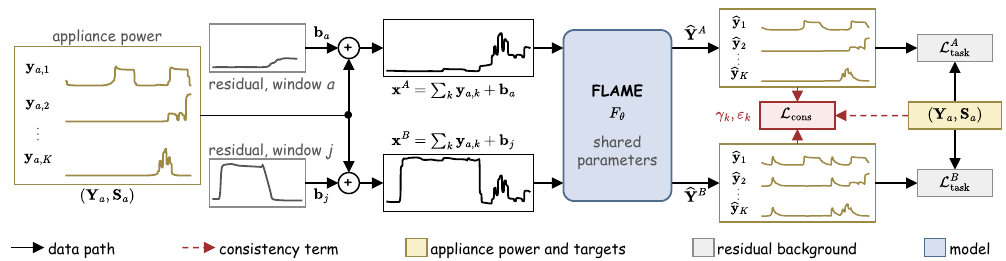}
  \caption{Label-preserving aggregate recomposition and training objective.
  The appliance power of window $a$ is added to its own residual background $\mathbf{b}_{a}$ and to that of another recorded window $j$, giving two aggregates that differ only in the residual background~\eqref{eq:background-recomposition} and share the targets $(\mathbf{Y}_{a},\mathbf{S}_{a})$; one model predicts both windows under the full task loss, and $\mathcal{L}_{\mathrm{cons}}$ uses predictions of the same pair from a second forward pass with dropout disabled.
  Traces: Reference Energy Disaggregation Data Set (REDD).}
  \label{fig:cbc-construction}
\end{figure*}

Both training constructions run on the same operation.
The residual background of a recorded window is separated by~\eqref{eq:additive-observation}, one side of that sum is replaced, and the aggregate is rebuilt through the same equation.
The additive relation therefore holds exactly on every constructed window.
Holding the residual background fixed and substituting target-appliance sequences produces further training windows, whose targets are recomputed from the substituted sequences.
Holding the target sequences fixed and substituting the residual background instead produces two windows that carry identical targets.
These are the pairs used for prediction consistency.

\subsubsection{Training-Window Synthesis}
\label{sec:window-synthesis}

Several target appliances operate for only a small fraction of the recording.
Windows drawn uniformly from the source pool are therefore dominated by intervals in which those appliances are off, leaving few examples of the behavior the model must learn to separate.
We synthesize training windows in the fixed-background direction of this construction: the residual background of a recorded window, termed the host window, is retained, and one target-appliance power sequence is substituted.

Appliance $k$ is treated as sparse when the fraction of samples in the training portion whose power exceeds its on power threshold $\tau_{k}$ falls below a sparsity threshold $\rho_{\mathrm{sp}}$.
This threshold governs which segments and windows the construction draws on; it is a per-appliance power level and is not the operating-state threshold that produces the state targets $\mathbf{s}_{k}$.
Each candidate window is labeled by the sparse appliance for which more than a minimum number of the window's samples exceed $\tau_{k}$; these samples need not be consecutive.
Every mini-batch is then filled to a fixed per-appliance quota rather than to a probability in expectation.
The appliance assigned to a position in the batch is resampled, over segments and scaling modes, until the synthesized window meets the same requirement for it.
The assigned count and the activation it stands for therefore coincide by construction.
The quota fixes how many positions are assigned to each appliance, not the number of windows in which that appliance is active.
Each position carries a single assignment; any other sparse appliance active in the host window is left as recorded.

Within a host window, the residual background $\mathbf{b}_i$ is held fixed.
The power sequence of the assigned appliance is substituted by an activation segment drawn from a per-appliance pool of segments detected on the training portion, and the aggregate is rebuilt through~\eqref{eq:additive-observation}.
The additive relation continues to hold exactly, and both targets follow from the substituted sequence.
Each segment may be rescaled in amplitude, in duration, or in both before insertion; duration scaling is implemented by interpolation, after which the segment is inserted with the partial-visibility semantics of~\cite{XiongEtAl2024MATNilm}.

\subsubsection{Label-Preserving Aggregate Recomposition}
\label{sec:background-recomposition}

\begin{figure*}[t]
  \centering
  \includegraphics[width=\textwidth]{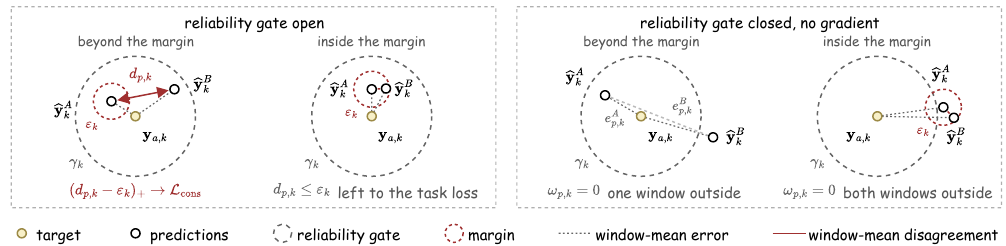}
  \caption{The consistency term drawn on the three window-mean quantities of~\eqref{eq:cbc-gate}--\eqref{eq:cbc-margin}, whose values the distances reproduce: the target is the center, the reliability gate is the circle of radius $\gamma_{k}$, which a pair passes only when both predictions lie inside it, and the disagreement is the side between the two predictions, penalized only beyond the margin $\varepsilon_{k}$.
  The four cases are the four combinations of the gate and the margin; only the first carries a penalty, and the gated excess is averaged over pairs and appliances~\eqref{eq:cbc-consistency}.}
  \label{fig:cbc-consistency}
\end{figure*}

The anchor of a pair is itself produced by the synthesis of Section~\ref{sec:window-synthesis}.
The anchor retains its host window's residual background, while the replacement residual background comes from another recorded source window.
Both are computed from the measured aggregate and target-appliance sequences of their source windows as in~\eqref{eq:additive-observation}, and the paired windows share the same synthesized target matrix.
Fig.~\ref{fig:cbc-construction} illustrates such a pair.

Constructing the pair requires a replacement residual background that differs from the anchor window's while leaving every modeled appliance target untouched.
For anchor window $a$, window $j$ is drawn from a pool of recorded source windows.

Using~\eqref{eq:additive-observation}, oriented pair $p=(a,j)$ is
\begin{equation}
  \begin{aligned}
    \mathbf{b}_{a}
      &=\mathbf{x}_{a}-\sum_{k=1}^{K}\mathbf{y}_{a,k},
    &\mathbf{b}_{j}
      &=\mathbf{x}_{j}-\sum_{k=1}^{K}\mathbf{y}_{j,k},\\
    \mathbf{x}_{p}^{A}
      &=\sum_{k=1}^{K}\mathbf{y}_{a,k}+\mathbf{b}_{a},
    &\mathbf{x}_{p}^{B}
      &=\sum_{k=1}^{K}\mathbf{y}_{a,k}+\mathbf{b}_{j},\\
    (\mathbf{Y}_{p}^{A},\mathbf{S}_{p}^{A})
      &=(\mathbf{Y}_{p}^{B},\mathbf{S}_{p}^{B})
       =(\mathbf{Y}_{a},\mathbf{S}_{a})
  \end{aligned}
  \label{eq:background-recomposition}
\end{equation}
where the superscripts $A$ and $B$ index the anchor and the recomposed window of the pair.
Every modeled appliance power sequence is preserved pointwise.
The operating-state targets are obtained from those sequences by thresholding, so they are preserved with them.
This label preservation is an arithmetic identity of the construction: the two aggregates differ exactly by $\mathbf{b}_{j}-\mathbf{b}_{a}$.

\emph{Admissibility.} The replacement residual background must be numerically valid, and the recomposed aggregate must stay within the configured physical range.
Candidates failing either check are resampled; if no valid replacement is found, sampling stops with an error.

\emph{Scope.} The recomposed window is obtained by label-preserving aggregate recomposition; it is not presented as a sample from the true joint distribution or as an identified causal intervention.

\subsection{Prediction Consistency}
\label{sec:cbc-framework}

Given the label-preserving pair of Section~\ref{sec:background-recomposition}, both windows receive the same complete power and operating-state task supervision.
The consistency term additionally penalizes disagreement between their corresponding per-appliance power predictions only when both predictions satisfy a fixed reliability criterion and only to the extent that their disagreement exceeds a fixed margin.
That penalty is placed on the gated power outputs of~\eqref{eq:representation-interface} rather than on the representations that produce them.
The two windows carry pointwise identical targets, so agreement between their predictions concerns the same quantity the task loss already scores.
Agreement between their representations would additionally require the two windows to be encoded identically despite their different residual backgrounds, which is more than the construction establishes.

\subsubsection{Reliability Gate and Margin}
\label{sec:cbc-scope}
\label{sec:cbc-gating}

Exact label preservation makes output consistency well defined, but it does not determine when or how strictly agreement should be enforced during training.
The consistency term therefore uses two appliance-specific thresholds fixed in advance: a reliability threshold $\gamma_{k}$ determines whether pair $p$ enters the term for appliance $k$, and a margin $\varepsilon_{k}$ determines how much disagreement the term tolerates.

The per-window task losses are evaluated on the ordinary training forward pass, including the configured dropout.
For the consistency term, the same paired batch is evaluated a second time with the dropout modules disabled while all other modules remain in their training states.
Throughout~\eqref{eq:cbc-gate}--\eqref{eq:cbc-consistency}, $\widehat{y}^{v}_{p,k,t}$ denotes a prediction from this dropout-disabled pass.
Both windows' predictions remain differentiable; only the binary gate defined below is detached.

Let $y_{p,k,t}$ denote the common target of appliance $k$ at time $t$ in pair $p$.
For each appliance, the reliability gate is computed from the window-mean absolute error of each window's prediction against that target,
\begin{equation}
  \begin{gathered}
    e^{v}_{p,k}=\frac{1}{T}\sum_{t=1}^{T}\left|\widehat{y}^{v}_{p,k,t}-y_{p,k,t}\right|,\\
    \omega_{p,k}=\mathbb{1}\!\left[e^{A}_{p,k}\le\gamma_{k}\right]\mathbb{1}\!\left[e^{B}_{p,k}\le\gamma_{k}\right]
  \end{gathered}
  \label{eq:cbc-gate}
\end{equation}
so the pair enters the term for appliance $k$ only when both windows are already predicted to within $\gamma_{k}$.
This excludes any pair in which either window is predicted unreliably, so the term never couples a reliable prediction with an unreliable one; it is also the only place where the target enters the consistency term.
The gate is evaluated without gradient.
It depends on the model's own prediction errors, so a gate that carried gradient would allow the term to be reduced by closing the gate rather than by reducing disagreement.

For a pair that passes the gate, the window-mean disagreement between the two predictions is defined as
\begin{equation}
  d_{p,k}=\frac{1}{T}\sum_{t=1}^{T}\left|\widehat{y}^{A}_{p,k,t}-\widehat{y}^{B}_{p,k,t}\right|
  \label{eq:cbc-margin}
\end{equation}
and the consistency term penalizes it only in excess of the margin, as $(d_{p,k}-\varepsilon_{k})_{+}$ with $(\cdot)_{+}=\max(\cdot,0)$.
Once disagreement is within the margin, the consistency penalty is zero and the task loss continues to fit each window to the common target.
Fig.~\ref{fig:cbc-consistency} draws the gate and the margin on these three window-mean quantities.

Both thresholds are set before training with the consistency term, from dropout-disabled predictions of checkpoints trained without it: $\gamma_{k}$ is a quantile of $\max(e^{A}_{p,k},e^{B}_{p,k})$ over source-side training pairs of those checkpoints, and $\varepsilon_{k}$ is a quantile of $d_{p,k}$ over the pairs that pass the resulting gate.
Setting them updates no model parameters, and they are held fixed throughout training.
Were they adapted while training, the term could shift toward pairs that are easier to satisfy, and any resulting change could not be attributed to a stated rule.

\subsubsection{Output Consistency}
\label{sec:cbc-loss}

Let $\mathcal{A}\subseteq\{1,\ldots,K\}$ denote the appliances to which the term is applied and $P$ the number of pairs in a mini-batch.
The consistency term averages the gated excess disagreement over pairs and then over appliances:
\begin{equation}
  \mathcal{L}_{\mathrm{cons}}
    =\frac{1}{|\mathcal{A}|}\sum_{k\in\mathcal{A}}
     \frac{1}{P}\sum_{p=1}^{P}
     \omega_{p,k}\,\bigl(d_{p,k}-\varepsilon_{k}\bigr)_{+}
  \label{eq:cbc-consistency}
\end{equation}
A pair closed by the gate contributes zero to the numerator and stays in the denominator, so the term scales with how often the gate opens instead of inflating when few pairs qualify.
Averaging within an appliance before combining appliances keeps an appliance with many gate-open pairs from dominating one with few.

Both windows carry the same targets and receive the same task loss, so $\mathcal{L}_{\mathrm{cons}}$ adds no target information.
Write $e^{v}_{p,k,t}=\widehat{y}^{v}_{p,k,t}-y_{p,k,t}$ for the pointwise prediction error against the common target; then $|\widehat{y}^{A}_{p,k,t}-\widehat{y}^{B}_{p,k,t}|=|e^{A}_{p,k,t}-e^{B}_{p,k,t}|$ at every position.
The disagreement $d_{p,k}$ is therefore the window-mean absolute difference between the two pointwise prediction errors and cancels any additive pointwise error component shared by both windows.
This cancellation applies to $d_{p,k}$, not to the full gated consistency loss, because the reliability gate depends on each window's mean absolute error.
The term is evaluated only on gate-open pairs and only in excess of the margin, but its gradient reaches the complete upstream computation that produces those predictions, including the shared stage of Section~\ref{sec:flame-stage1}.
The gate and the margin therefore determine whether the term is active, not which appliances or parameters the resulting updates can affect.

\subsection{Overall Training Objective}
\label{sec:joint-optimization}

For $v\in\{A,B\}$, let $\widehat{\mathbf{R}}^{v}$, $\widehat{\boldsymbol{\ell}}^{v}$, and $\widehat{\mathbf{Y}}^{v}$ stack the raw regression outputs, state logits, and gated power outputs of~\eqref{eq:representation-interface} across appliances and time.
Using the common anchor targets $(\mathbf{Y},\mathbf{S})$, the task and total objectives are
\begin{equation}
  \begin{aligned}
    \mathcal{L}_{\mathrm{task}}^{v}
      ={}&\alpha_{\mathrm{reg}}
        \operatorname{MSE}(\widehat{\mathbf{R}}^{v},\mathbf{Y})
        +\alpha_{\mathrm{state}}
        \operatorname{BCE}(\widehat{\boldsymbol{\ell}}^{v},\mathbf{S})\\
       &+\alpha_{\mathrm{gate}}
        \operatorname{MSE}(\widehat{\mathbf{Y}}^{v},\mathbf{Y}),\\
    \mathcal{L}
      ={}&\tfrac{1}{2}\left(
        \mathcal{L}_{\mathrm{task}}^{A}
        +\mathcal{L}_{\mathrm{task}}^{B}\right)
        +\lambda_{\mathrm{cons}}\mathcal{L}_{\mathrm{cons}}
  \end{aligned}
  \label{eq:joint-objective}
\end{equation}
where $\operatorname{MSE}$ denotes the mean squared error and $\operatorname{BCE}$ the binary cross-entropy evaluated from logits.
Every component of the per-window task loss is applied to both windows, and the two per-window task losses are weighted equally.
At each update, the task-loss gradient is computed first, the dropout-disabled consistency pass is then evaluated, and its gradient is accumulated into the same parameter gradients before one clipping operation and one optimizer step.
This sequential accumulation implements the gradient sum in~\eqref{eq:joint-objective} without retaining both computation graphs simultaneously.
The loss weights and optimizer settings are reported with the experimental settings.

Training uses the recomposed window, the two fixed thresholds, and $\mathcal{L}_{\mathrm{cons}}$; inference uses none of them and runs the standard single-window forward path.
Relative to the same checkpoint, the proposed method therefore introduces no additional inference-time module or parameter.

\section{Experiments and Results}
\label{sec:experiments-results}

\begin{table*}[t]
  \caption{Results on the unseen test household of REDD, UK-DALE, and REFIT. DW, FR, MW, WD, KT, and WM denote the dishwasher, refrigerator, microwave, washer--dryer, kettle, and washing machine; Ave is the unweighted mean over the appliances of each data set. Proposed denotes the complete method trained on label-preserving pairs. Bold marks the better value.}
  \label{tab:redd-main-results}
  \centering
  \scriptsize
  \setlength{\tabcolsep}{3pt}
  \renewcommand{\arraystretch}{1.1}
  \begin{tabular}{@{}llccccc|cccccc|cccccc@{}}
    \toprule
    & & \multicolumn{5}{c|}{REDD} & \multicolumn{6}{c|}{UK-DALE} & \multicolumn{6}{c}{REFIT} \\
    Metric & Model & DW & FR & MW & WD & Ave & KT & MW & FR & DW & WM & Ave & KT & MW & FR & DW & WM & Ave \\
    \midrule
    MAE (W) & Single-window FLAME & 12.33 & 22.77 & 14.70 & 9.18 & 14.75 & \textbf{3.85} & 3.23 & \textbf{16.32} & 16.18 & 4.81 & 8.88 & \textbf{16.19} & \textbf{2.69} & 26.08 & 25.11 & \textbf{9.07} & 15.83 \\
            & Proposed & \textbf{10.22} & \textbf{20.48} & \textbf{14.45} & \textbf{7.40} & \textbf{13.14} & 4.37 & \textbf{2.90} & 17.47 & \textbf{14.34} & \textbf{3.46} & \textbf{8.51} & 16.80 & 2.71 & \textbf{22.18} & \textbf{21.78} & 9.28 & \textbf{14.55} \\
    \midrule
    SAE (W) & Single-window FLAME & 11.85 & 16.76 & 13.35 & 7.90 & 12.46 & \textbf{3.24} & 2.94 & \textbf{11.67} & 14.94 & \textbf{3.30} & 7.22 & \textbf{15.08} & \textbf{2.49} & 20.29 & 21.20 & \textbf{6.11} & 13.03 \\
            & Proposed & \textbf{9.49} & \textbf{14.58} & \textbf{12.77} & \textbf{5.72} & \textbf{10.64} & 3.79 & \textbf{2.65} & 13.10 & \textbf{12.89} & \textbf{1.84} & \textbf{6.85} & 16.05 & 2.49 & \textbf{17.93} & \textbf{17.03} & 6.70 & \textbf{12.04} \\
    \midrule
    F1      & Single-window FLAME & 0.75 & \textbf{0.86} & \textbf{0.53} & \textbf{0.91} & 0.76 & 0.98 & 0.42 & \textbf{0.72} & \textbf{0.75} & 0.80 & 0.73 & \textbf{0.70} & 0.26 & 0.77 & 0.73 & 0.79 & 0.65 \\
            & Proposed & \textbf{0.78} & 0.86 & 0.52 & 0.91 & \textbf{0.76} & \textbf{0.98} & \textbf{0.43} & 0.71 & 0.74 & \textbf{0.86} & \textbf{0.74} & 0.62 & \textbf{0.27} & \textbf{0.82} & \textbf{0.77} & \textbf{0.80} & \textbf{0.66} \\
    \bottomrule
  \end{tabular}
\end{table*}

We evaluate the complete method of Section~\ref{sec:proposed-method} against single-window FLAME on three public low-frequency data sets, and compare the FLAME architecture with four alternative multi-appliance architectures under a common training protocol.

\subsection{Datasets and Evaluation Metrics}
\label{sec:evaluation-protocol}

Experiments are conducted on REDD~\cite{KolterJohnson2011REDD}, UK-DALE~\cite{KellyKnottenbelt2015UKDALE}, and REFIT~\cite{MurrayEtAl2017REFIT}.
On REDD we jointly model the dishwasher, refrigerator, microwave, and washer--dryer; on UK-DALE and REFIT the kettle, microwave, refrigerator, dishwasher, and washing machine.
REDD and UK-DALE are used at their 6-s resolution and REFIT at 8~s; all power channels are scaled by a constant of 612~W.
Training and testing use disjoint households: House~3 of REDD, House~1 of UK-DALE, and Houses~3, 5, 9, and~20 of REFIT are used for training, and House~1 of REDD and House~2 of UK-DALE and REFIT for testing.
Each training recording is split chronologically into a training portion and a validation portion (the final 30\% on REDD and 5\% on UK-DALE and REFIT) before windowing, and the test household contributes no training data.
Target-appliance segments for training-window synthesis are drawn from additional training-only households (Houses~2, 4, 5, and~6 of REDD, House~5 of UK-DALE, and fifteen further houses of REFIT).
Signals are divided into 720-sample windows with a stride of 120 samples.

Three metrics are computed for each appliance $k$ over the $N_{\mathrm{w}}$ test windows: mean absolute error (MAE), signal aggregate error (SAE), and the F1 score of the operating state,
\begin{equation}
  \begin{gathered}
    \mathrm{MAE}_{k}=\frac{1}{N_{\mathrm{w}}T}\sum_{i=1}^{N_{\mathrm{w}}}\sum_{t=1}^{T}\left|\widehat{y}_{i,k,t}-y_{i,k,t}\right|,\\
    \mathrm{SAE}_{k}=\frac{1}{N_{\mathrm{w}}T}\sum_{i=1}^{N_{\mathrm{w}}}\left|\sum_{t=1}^{T}\widehat{y}_{i,k,t}-\sum_{t=1}^{T}y_{i,k,t}\right|,\\
    \mathrm{F1}_{k}=\frac{2P_{k}R_{k}}{P_{k}+R_{k}}
  \end{gathered}
  \label{eq:metrics}
\end{equation}
where $\widehat{y}_{i,k,t}$ and $y_{i,k,t}$ are the predicted and target power of appliance $k$ at sample $t$ of test window $i$, both in watts, and $P_{k}$ and $R_{k}$ are the precision and recall of the predicted operating state $\widehat{s}_{i,k,t}=\mathbb{1}\!\left[\sigma(\widehat{\ell}_{i,k,t})>0.3\right]$ against the operating-state target $s_{i,k,t}$ over all test-window samples, with 0.3 the state-probability threshold.
Macro values are unweighted means over appliances.

\subsection{Experimental Setup}
\label{sec:cbc-training-setup}

We compare two variants.
The baseline, single-window FLAME, trains FLAME on single windows with $\mathcal{L}_{\mathrm{task}}$ alone.
The proposed method constructs the label-preserving pair, applies the complete task loss to both windows, evaluates the consistency term on the dropout-disabled pass of Section~\ref{sec:cbc-gating}, and sets $\lambda_{\mathrm{cons}}=0.8$.
The comparison therefore evaluates the proposed method as a whole rather than attributing its result to an individual component.
All reported values are averaged over three independent trials; for macro MAE we also report the sample standard deviation and the number of paired runs in which the proposed method has lower error than single-window FLAME.

Both variants use the same FLAME configuration: a ResTCN backbone of width 256 and depth 4 (parallel kernels 3/9/3, block dilations 7/11/17/23), $N=2$ routing levels, a shared TConv expert with kernel 3, per-appliance BiGRU (hidden size 128) and BiMamba (state size 16) experts, SE gates without channel reduction, gate MLPs of width 128, and dropout 0.1.
They are trained separately from scratch with AdamW (learning rate $10^{-3}$, weight decay $10^{-2}$) for 45 epochs of 30 updates on REDD, 90 epochs of 50 updates on UK-DALE, and 60 epochs of 50 updates on REFIT; single-window FLAME uses 64 windows per batch, and the proposed method uses 64 pairs, or 128 windows, with complete power and state supervision on both members of every pair.
Operating-state targets use a 10-W threshold, and the task-loss weights are $(\alpha_{\mathrm{reg}},\alpha_{\mathrm{state}},\alpha_{\mathrm{gate}})=(2,1,1)$.

For training-window synthesis (Section~\ref{sec:window-synthesis}), activation segments are detected on the training portion before windowing with per-appliance on power thresholds $\tau_{k}$, and the sparsity threshold is $\rho_{\mathrm{sp}}=0.2$.
For each paired example, the replacement residual background is taken from another recorded window.
The consistency set $\mathcal{A}$ contains the dishwasher, microwave, and washer--dryer on REDD; the kettle, microwave, refrigerator, dishwasher, and washing machine on UK-DALE; and the kettle, microwave, dishwasher, and washing machine on REFIT.
Appliances outside $\mathcal{A}$ remain fully supervised in both windows but have zero weight in $\mathcal{L}_{\mathrm{cons}}$.

A separate experiment varies the architecture instead of the training method.
Five multi-appliance architectures---HMMOE~\cite{LiangEtAl2026HMMOE}, MATNilm~\cite{XiongEtAl2024MATNilm}, BERT4NILM~\cite{YueEtAl2020BERT4NILM}, MA-NILM~\cite{LiEtAl2026MAFeatures}, and FLAME---are trained and evaluated on each of the three data sets.
All five follow one common protocol with the same households, windowing, task loss, optimizer settings, per-data-set training budget, and inference procedure; no pair is constructed, the consistency term is not applied, and no hyperparameter is tuned for an individual architecture.
BERT4NILM is trained directly under this protocol and does not reproduce the masked training of the original model, and MA-NILM is a sequence-to-sequence adaptation of a sequence-to-point design.

\subsection{Experimental Results}
\label{sec:redd-main-results}

Table~\ref{tab:redd-main-results} reports MAE, SAE, and F1 on the unseen test household of each data set.
The proposed method attains lower macro MAE and SAE than single-window FLAME on all three data sets.
On REDD, macro MAE falls from $14.75\pm0.30$ to $13.14\pm0.50$~W, a reduction of 1.61~W (10.9\%) with lower error in three of three paired runs and a paired 95\% confidence interval of $[-2.59,-0.62]$~W, and macro SAE falls from 12.46 to 10.64~W.
On REFIT, macro MAE falls from $15.83\pm0.62$ to $14.55\pm0.66$~W (1.28~W, 8.1\%; three of three runs) and macro SAE from 13.03 to 12.04~W.
On UK-DALE, macro MAE falls from $8.88\pm1.53$ to $8.51\pm0.36$~W (0.37~W, 4.2\%) and macro SAE from 7.22 to 6.85~W; on UK-DALE and REFIT the paired 95\% confidence intervals of the macro differences include zero.

\subsection{Architecture Comparison}
\label{sec:architecture-comparison}

\begin{table}[t]
  \caption{Macro MAE (W) of five architectures under the common training protocol, as mean $\pm$ sample standard deviation over three independent trials. Bold marks the lowest value on each data set.}
  \label{tab:architecture-comparison}
  \centering
  \footnotesize
  \setlength{\tabcolsep}{5pt}
  \renewcommand{\arraystretch}{1.1}
  \begin{tabular}{@{}lccc@{}}
    \toprule
    Architecture & REDD & UK-DALE & REFIT \\
    \midrule
    HMMOE           & 15.98 $\pm$ 0.22 & 10.88 $\pm$ 0.82 & 19.78 $\pm$ 2.11 \\
    MATNilm         & 17.04 $\pm$ 1.79 & 10.23 $\pm$ 1.04 & 18.75 $\pm$ 0.26 \\
    BERT4NILM       & 16.70 $\pm$ 0.96 & 13.60 $\pm$ 1.31 & 23.53 $\pm$ 1.03 \\
    MA-NILM         & 26.74 $\pm$ 0.77 & 14.63 $\pm$ 0.78 & 22.87 $\pm$ 0.78 \\
    FLAME           & \textbf{14.07 $\pm$ 0.91} & \textbf{8.30 $\pm$ 0.25} & \textbf{14.57 $\pm$ 0.24} \\
    \bottomrule
  \end{tabular}
\end{table}

Table~\ref{tab:architecture-comparison} reports the macro MAE of the five architectures on the unseen test household of each data set.
FLAME attains the lowest macro MAE among the compared architectures under this common protocol on all three data sets.
Because the four alternatives are reimplemented under this protocol without architecture-specific tuning, the table compares them under a single shared optimization setting rather than at the best configuration reported for each original method; the ranking may therefore reflect both architectural properties and compatibility with that setting.

\section{Conclusion}
\label{sec:conclusion}

This paper introduced a method for multi-appliance NILM that combines label-preserving aggregate recomposition with prediction consistency during training.
The construction uses the additive decomposition of aggregate power and time-aligned submetered measurements to recompose an aggregate window so that only its residual background changes while every modeled appliance power and operating-state target is preserved pointwise.
Training then constrains the relation between the per-appliance power predictions of the two windows under complete task supervision.
The consistency term acts only on pairs that pass a reliability gate and only on disagreement beyond a margin; both thresholds are fixed in advance and together determine whether the term is active, without restricting which parameters its gradient can reach.
The proposed method was implemented using FLAME, a multi-appliance architecture with two-stage shared-to-specific expert routing.
At deployment the model runs a standard single-window forward pass, so the proposed method adds no inference-time module or parameter.

On REDD, UK-DALE, and REFIT, each evaluated on an unseen household, the proposed method with $\lambda_{\mathrm{cons}}=0.8$ lowers macro MAE from 14.75 to 13.14~W, from 8.88 to 8.51~W, and from 15.83 to 14.55~W relative to single-window training.
In a separate comparison under a common training protocol that varies only the architecture, FLAME attains the lowest macro MAE among the compared architectures on all three data sets.

The scope of the method-comparison evidence is set by the protocol that produced it: three data sets, one unseen household each, and one fixed consistency weight.
The recomposed window is label-preserving by construction, a property of that construction rather than a claim about the true joint distribution or a physically realized intervention.

Future work will extend the evaluation to further data sets and household protocols and examine how the fixed gate and margin shape performance across appliance conditions.

\bibliographystyle{IEEEtran}
\bibliography{refs}

\end{document}